\documentclass[submission,copyright,creativecommons]{eptcs}
\providecommand{\event}{ICLP 2020} 
\usepackage{breakurl}             
\usepackage{underscore}           
\usepackage{float}
\usepackage{amssymb}
\usepackage{amsmath}
\usepackage{url}
\usepackage{color}
\usepackage{graphicx}
\usepackage{booktabs}
\usepackage[normalem]{ulem}

\usepackage{listings}
\usepackage{graphicx}

\usepackage{float}

\usepackage{hyperref}

\usepackage[ruled,vlined]{algorithm2e}

\usepackage{IEEEtrantools}

\newtheorem{example}{Example}

\title{A System for Explainable Answer Set Programming\thanks{Partially supported by MINECO, Spain, grant TIN2017-84453-P and CITIC Research Center, Xunta de Galicia, Spain and ERDF (ED431G 2019/01). The second author is funded by the Alexander von Humboldt Foundation, Germany.}}

\author{Pedro Cabalar
\institute{University of Corunna, Spain}
\email{cabalar@udc.es}
\and
Jorge Fandinno
\institute{University of Potsdam, Germany}
\email{fandinno@uni-potsdam.de}
\and
Brais Mu\~niz*
\institute{University of Corunna, Spain}
\email{brais.mcastro@udc.es}
}

\def\titlerunning{A System for Explainable Answer Set Programming}
\def\authorrunning{P. Cabalar, J. Fandinno, B. Mu\~niz}

\def\xclingo{{\tt  xclingo}}
\newcommand{\tracerule}{\texttt{trace\_rule}}
\newcommand{\trace}{\texttt{trace}}
\newcommand{\showtrace}{\texttt{show\_trace}}

\newcommand\keywordsname{KEYWORDS}
\newenvironment{keywords}
  {\noindent\normalfont\small\rmfamily{\em \keywordsname}:}%
  \endlist 

\begin{document}
\maketitle

\begin{abstract}
We present \xclingo{}, a tool for generating explanations from ASP programs annotated with text and labels.
These annotations allow tracing the application of rules or the atoms derived by them.
The input of \xclingo{} is a markup language written as ASP comment lines, so the programs annotated in this way can still be accepted by a standard ASP solver.
\xclingo{} translates the annotations into additional predicates and rules and uses the ASP solver {\tt clingo} to obtain the extension of those auxiliary predicates.
This information is used afterwards to construct derivation trees containing textual explanations.
The language allows selecting which atoms to explain and, in its turn, which atoms or rules to include in those explanations.
We illustrate the basic features through a diagnosis problem from the literature.
\end{abstract}


\section{Introduction}
\label{sec:intro}



Answer Set Programming (ASP)~\cite{Nie99,MT99,brewka2011} is a successful paradigm for Knowledge Representation and problem solving. Under this paradigm, the programmer represents a problem as a logic program formed by a set of rules and obtains solutions to that problem in terms of models of the program called answer sets. Thanks to the availability of efficient solvers, ASP is nowadays applied in a wide variety of areas including robotics, bioinformatics, music composition~\cite{housekeeping-robotics-Erdem12,phylogenetic-trees-Brooks07,Music-Boenn2010}, and many more. 

An ASP program does not contain information about the method to obtain the answer sets, something that is completely delegated to the ASP solver.
This, of course, has the advantage of making ASP a fully declarative language, where the programmer must concentrate on specification rather than on design of search algorithms.
However, when it comes to \emph{explainability} of the obtained results, the information provided by answer sets themselves is usually scarce.
%
%
There exist several approaches for obtaining justifications for answer sets: for a recent review, see \cite{FandinnoS19}. Some of them are more oriented to \textit{debugging of ASP programs} while others are interested in the causal nature of justifications themselves. What these approaches generally do is to offer some kind of enlightening about the derivation process of the rules that led to finally include (or not) some literal in an answer set.

Justifying the result of an ASP program is not only interesting for the programmer but has also other implications, especially in the context of \emph{explainable Artificial Intelligence}. For instance, since the approval of the General Data Protection Regulation (GDPR) by the European Union every system that makes automatic decisions that affect to persons must offer some kind of \emph{explanation} on the logic involved in the decision making process, obviously in a human-readable way.

In this paper, we present \xclingo{}, a tool for generating explanations of annotated programs for the ASP solver {\tt clingo}~\cite{gekakaosscwa16a}.
The input accepted by \xclingo{} is a markup language that introduces annotations through program comments, specifying which rules or atoms must be traced.
Using these annotations, \xclingo{} generates derivation trees following the framework of \textit{causal graph justifications} introduced in~\cite{CabalarFF14cg}.
Under that framework, answer sets are multi-valued interpretations where the value for each true atom is an algebraic expression constructed with labels associated to program rules. Each expression is an alternative of multiple derivation trees that have been proved to correspond to the set of minimal proofs for the atom built with the Horn clauses in the program reduct. By now, \xclingo{} just guarantees correctness of the obtained derivations, but not their minimality, which is planned to become a future optional feature.
The tool uses the \texttt{clingo} python API to translate the annotations into additional rules with auxiliary predicates and computes the explanations from the information obtained from these predicates.

Since causal graphs show possible derivations for the conclusions, it is difficult to avoid that, as the size of a rule system increases, the readability and comprehension of the explanations becomes more difficult.
A large explanation may be tractable by a computer but not too useful for a human reader, who is normally more concerned about \emph{relevant} pieces of information. 
Because of that, \xclingo{} puts an extra effort on creating a flexible way to format these explanations at a detailed level, style and size. 

Although its main purpose is the explanation of ASP program conclusions, we also find \xclingo{} helpful for program debugging, again, because of the flexibility of its explanation configuration system.

In order to show its features and to demonstrate its usefulness, we use a diagnosis problem example from the literature as a guide. We choose this example because we find the nature of \xclingo{}'s explanations very helpful in diagnosis.

The rest of the paper is organised as follows. First we introduce our diagnosis running example. Then, the input language and the features of \xclingo{} are described. Next, we describe how the translation of the input into a standard ASP program works. Afterwards, we describe how the explanations are computed from the solution of the translated program. Finally, we comment about related work and we conclude the paper.


\section{Motivating Example}
\label{sec:motivating-example}

\begin{figure}[htbp]
  \centering
  \includegraphics[scale=0.7]{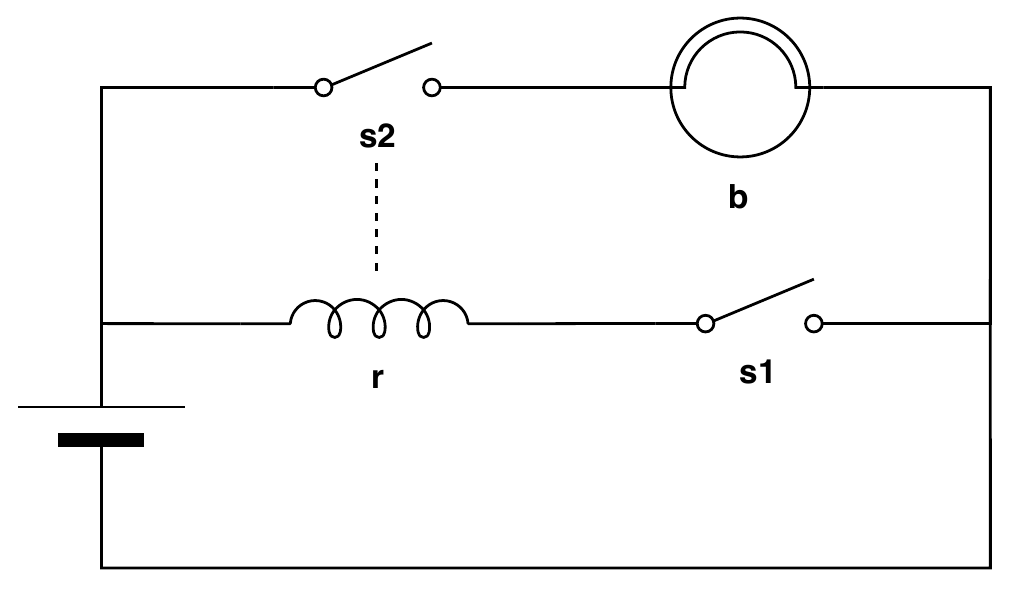}
  \caption{A circuit with a bulb $b$, a relay $r$ and two switches, $s1$ and $s2$.}
  \label{fig:1}
\end{figure}

%
We consider an example from \cite{gelfond2003diagnosis} (Fig.~\ref{fig:1}).
In the example, an analog $AC$ circuit is presented. In it, an agent can close a switch that should ultimately cause a bulb to turn on. 
However, there are exogenous actions that can modify the environment and make the bulb not to turn on by closing the switch. 
Our goal is to develop a diagnostic system that can identify the reasons why the light does not turn on and present them to the user in the form of readable explanations.

\begin{example}
\label{ex:1}
\mbox{\bf (From Balduccini and Geldfond 2003)}
Consider a system $S$ consisting of an agent operating an analog circuit $AC$ from Fig.~\ref{fig:1}. We assume that switches $s_{1}$ and $s_{2}$ are mechanical components which cannot become damaged. Relay $r$ is a magnetic coil. If not damaged, it is activated when $s_{1}$ is closed, causing $s_{2}$ to close. Undamaged bulb $b$ emits light if $s_{2}$ is closed. For simplicity of presentation we consider the agent capable of performing only one action, $close(s_{1})$. The environment can be represented by two damaging exogenous$^{1}$ actions: $brk$, which causes $b$ to become faulty, and $srg$ (power surge), which damages $r$ and also $b$ assuming that $b$ is not protected. Suppose that the agent operating this device is given a goal of lighting the bulb. He realizes that this can be achieved by closing the first switch, performs the operation, and discovers that the bulb is not lit.
\end{example}

Our ASP implementation of this example (we call program $P_1$ in  Listings~\ref{listing:p1a} and \ref{listing:p1b}) follows the one presented in \cite{gelfond2003diagnosis} with the addition of {\tt c/3} and few other predicates for improving the explanation results.
At a first sight, the encoding may seem too involved for our small example, but this is because the representation is general enough to  cover a whole family of similar diagnosis problems.
Listing~\ref{listing:p1a} contains the basic type definitions.
The predicate names are self-explanatory, except perhaps lines 15--24.
This is because we allow arbitrary fluent domains that can be specified explicitly through predicate {\tt value(F,V)}, meaning that fluent {\tt F} may have value {\tt V}.
When no value has been specified in that way, fluents are assumed Boolean by default.
Finally, predicate {\tt domain(F,V)} collects all domain values for {\tt V}, regardless of whether they are defined explicitly or by default.
In our example, fluents {\tt relay, light, s1} and {\tt s2} can take values {\tt on} and {\tt off} (to make them more readable) whereas the rest of fluents are Boolean.

\begin{figure}[htbp]
\begin{lstlisting}[caption={Type predicates for program $P_1$.},label={listing:p1a}]
plength(1).
time(0..L) :- plength(L).
step(1..L) :- plength(L).

switch(s1). switch(s2).
component(relay). component(bulb).

fluent(relay).
fluent(light).
fluent(b_prot).
fluent(S):-switch(S).
abfluent(ab(C)) :- component(C).
fluent(F) :- abfluent(F).

value(relay,on). value(relay,off).
value(light,on). value(light,off).
value(S,open) :- switch(S).
value(S,closed) :- switch(S).
hasvalue(F) :- value(F,V).
% Fluents are boolean by default
domain(F,true) :- fluent(F), not hasvalue(F).
domain(F,false) :- fluent(F), not hasvalue(F).
% otherwise, they take the specified values
domain(F,V) :- value(F,V).


agent(close(s1)).
exog(break).
exog(surge).
action(Y):-exog(Y).
action(Y):-agent(Y).

\end{lstlisting}
\end{figure}

Listing~\ref{listing:p1b} contains the description of the problem.
Given any action {\tt A}, fluent {\tt F}, value {\tt V} and time point {\tt I} we use the following predicates:\\[5pt]
\begin{center}
\begin{tabular}{ll}
{\tt h(F,V,I)} & = {\tt F} holds value {\tt V} at {\tt I}\\
{\tt obs\_h(F,V,I)} & = {\tt F} was observed to hold value {\tt V} at {\tt I}\\
{\tt c(F,V,I)} & = {\tt F}'s value was caused to be {\tt V} at {\tt I}\\
{\tt c(F,I)} & = {\tt F}'s value was caused at {\tt I}\\
{\tt o(A,I)} & = {\tt A} occurred at {\tt I}\\
{\tt obs\_o(A,I)} & = {\tt A} was observed to occur at {\tt I}
\end{tabular}
\vspace{5pt}
\end{center}

\begin{figure}[htbp]
\begin{lstlisting}[caption={Program $P_1$ describing Example~\ref{ex:1}.}, label={listing:p1b}]
% Inertia
h(F,V,I) :- h(F,V,I-1), not c(F,I), step(I).

% Axioms for caused
h(F,V,J) :- c(F,V,J).
c(F,J)   :- c(F,V,J).

% Direct effects
c(s1,closed,I) :- o(close(s1),I), step(I).

% Indirect effects
c(relay,on,J)   :- h(s1,closed,J), h(ab(relay),false,J), time(J).
c(relay,off,J)  :- h(s1,open,J), time(J).
c(relay,off,J)  :- h(ab(relay),true,J), time(J).
c(s2,closed,J)  :- h(relay,on,J), time(J).
c(light,on,J)    :- h(s2,closed,J), h(ab(bulb),false,J), time(J).
c(light,off,J)   :- h(s2,open,J), time(J).
c(light,off,J)   :- h(ab(bulb),true,J), time(J).

% Malfunctioning
c(ab(bulb),true,I) :- o(break,I), step(I).
c(ab(relay),true,I) :- o(surge,I), step(I).
c(ab(bulb),true,I) :- o(surge,I), not h(b_prot,true,I-1), step(I).


% Executability
:- o(close(S),I), h(S,closed,I-1), step(I).

% Something happening actually occurs
o(A,I) :- obs_o(A,I), step(I).

% Check that observations hold
:- obs_h(F,V,J), not h(F,V,J).

% Completing the initial state
h(F,V,0) :- domain(F,V), not -h(F,V,0).
-h(F,V,0) :- h(F,W,0), domain(F,V), W!=V.


% A history
obs_h(s1,open,0).
obs_h(s2,open,0).
obs_h(b_prot,true,0).
obs_h(ab(bulb),false,0).
obs_h(ab(relay),false,0).
obs_o(close(s1),1).

% Something went wrong
obs_h(light,off,1).

% Diagnostic module: generate exogenous actions
o(Z,I) :- step(I), exog(Z), not no(Z,I).
no(Z,I) :- step(I), exog(Z), not o(Z,I).
\end{lstlisting}
\end{figure}
As usual in diagnosis problems, we differentiate between what happens in the real world, with predicates {\tt h/3} and {\tt o/2}, and the  partial observations we have about that world, with predicates {\tt obs\_h/3} and {\tt obs\_o/2}, respectively. If we execute {\tt clingo} on this code, we obtain the three answer sets that correspond to the possible diagnosis: one including an exogenous action  {\tt o(break,1)}; a second one with the exogenous action {\tt o(surge,1)}; and, finally, a third, non-minimal diagnosis where both exogenous actions occur.
Of course, in the original work by Balduccini and Gelfond, diagnoses were additionally minimised to avoid unnecessary addition of exogenous actions, but for the purpose of this paper, we consider the three answer sets of program $P_1$ as equally interesting for generating explanations.
At the moment, \xclingo{} cannot properly deal with minimisation clauses in {\tt clingo} yet.

In the rest of the paper, we use this code as a running example. We will complete it using different \xclingo{} features in order to get the diagnoses in a fully readable and understandable way.


\section{The \xclingo{} system}
\label{sec:language}

To understand the purpose of the different types of annotations in \xclingo{} it is perhaps better to start illustrating the kind of output we expect to achieve.
For instance, Fig.~\ref{fig:p1expl} shows the result we obtain for program $P_1$ encoding from Example~\ref{ex:1} once it is annotated --- we will see how these annotations look like later on.
As we can see, the programmer has requested explanations for fluents {\tt light} and {\tt relay}.
Each explanation must be understood as follows.
Below each explained (true) atom (preceded by {\tt >>}) we get a list of trees that correspond to alternative (and equally effective) causes for the atom.
Then, in each tree, two explanations at the same level must be understood as a joint cause (the two effects act together) and each time we jump into a lower level, we can read this as a ``because'' relation.

Fig.~\ref{fig:p1expl} shows the same three answer sets we obtain with plain {\tt clingo}.
Answer set $1$ corresponds to the case in which both a power surge occurred and something broke the bulb.
The explanation for the relay being off (lines 2--6) can be read as follows: ``the relay is not working at $1$ \emph{because} it has been damaged \emph{because} there has been a power surge.'' We have started all explanations for exogenous actions with the word {\tt Hypothesis} to clarify that these are assumptions added to explain the observations.
As we can see, in this first answer set, there are two alternative valid causes for the light being off.
The first one (Fig.~\ref{fig:p1expl}, lines 9--12) is that the bulb was damaged because something broke it.
%
The three lines in the explanation respectively come from the annotations (we will see later) for lines 18, 21 and 52 in Listing~\ref{listing:p1b}.
The second cause (lines 14--16) is that the light was already off in the initial state because switch {\tt s2} was initially open:
in this case, because of the activation of rules in lines 17 and 5 in Listing~\ref{listing:p1b}.
%
%
%

\begin{figure}[htbp]
\begin{lstlisting}
Answer: 1
>> h(relay,off,1)	[1]
  *
  |__"The relay is not working at 1"
  |  |__"The relay has been damaged at 1"
  |  |  |__"Hypothesis: there has been a power surge at 1"

>> h(light,off,1)	[2]
  *
  |__"The light is off at 1"
  |  |__"The bulb has been damaged at 1"
  |  |  |__"Hypothesis: something has broken the bulb at 1"

  *
  |__"The light is off at 1"
  |  |__"s2 was initially open"


Answer: 2
>> h(relay,off,1)	[1]
  *
  |__"The relay is not working at 1"
  |  |__"The relay has been damaged at 1"
  |  |  |__"Hypothesis: there has been a power surge at 1"

>> h(light,off,1)	[1]
  *
  |__"The light is off at 1"
  |  |__"s2 was initially open"


Answer: 3
>> h(relay,on,1)	[1]
  *
  |__"The relay is working at 1"
  |  |__"The agent has closed switch s1 at 1"
  |  |__"Initially, the relay was not damaged"

>> h(light,off,1)	[1]
  *
  |__"The light is off at 1"
  |  |__"The bulb has been damaged at 1"
  |  |  |__"Hypothesis: something has broken the bulb at 1"
\end{lstlisting}
\caption{Explanations obtained for the annotated version of $P_1$.}
\label{fig:p1expl}
\end{figure}

Answer set $2$ (lines 19--29) corresponds to the case in which we just had a power surge.
When this happens, the relay is not working (as in the answer set $1$) and the light simply remains off, since {\tt s2} was initially open.
In this case, we do not get the additional reason for having the light off, since the bulb is not broken.

Finally, answer set $3$ shows the case where something breaks the bulb but there is no power surge. 
In this case, we can see that the relay eventually worked because the agent closed switch $1$ and the relay was not initially damaged.
As nothing else happens, the relay is still undamaged in state $1$.
Notice that these two things (the agent closing the switch and the relay being initially undamaged) constitute a {\em joint cause} altogether: lines 36 and 37 share the same parent at the explanation tree.

Let us proceed now to describe how these explanations are generated, including the markup language accepted by \xclingo{}.
Each explanation is a tree that shows the derivation proof for an atom. In these trees, each node is a \emph{trace label} that corresponds to a \texttt{string} associated to some fired rule or to some derived atom.
Trace labels can be created manually through \textit{annotations} or can be automatically generated by \xclingo{}. 
For instance, Fig.~\ref{fig:auto-tracing} displays the explanation for \texttt{h(light,off,1)} under the ``auto-tracing'' mode, where every rule is automatically traced using the rule head as a label.
As we can see, we obtain a complete derivation tree for \texttt{h(light,off,1)} following the positive part of the program.
\begin{figure}[htbp]
\begin{lstlisting}
>> h(light,off,1)	[1]
  *
  |__h(light,off,1)
  |  |__c(light,off,1)
  |  |  |__h(ab(bulb),true,1)
  |  |  |  |__c(ab(bulb),true,1)
  |  |  |  |  |__o(break,1)
  |  |  |  |  |  |__step(1)
  |  |  |  |  |  |  |__plength(1)
  |  |  |  |  |  |__exog(break)
  |  |  |  |  |__step(1)
  |  |  |  |  |  |__plength(1)
  |  |  |__time(1)
  |  |  |  |__plength(1)
\end{lstlisting}
\caption{Explanation using automatically generated trace labels.}
\label{fig:auto-tracing}
\end{figure}
Note the difference with respect to Fig.~\ref{fig:p1expl}, where we used manually defined (textual) trace labels.
In that case, the tool skips any intermediate node in the derivation tree that has no explicit trace label.
In this way, knowledge engineers may decide the detail to be shown: either automatically tracing all possible rule applications, which may be helpful for debugging, or selecting the relevant information, something more interesting for explanation. 
In the latter, the explanation design, that is, selecting the right amount of information, may become a non-trivial Knowledge Representation effort in itself, but would easily become a problem instead if the tool did not offer such a possibility.
For instance, one important decision is to avoid tracing the inertia rule (line 2 of Listing~\ref{listing:p1b}). In that way, if we ask for the explanation of {\tt h(light,off,20)} in answer set 2 of Fig.~\ref{fig:p1expl} we still get the same derivation tree (replacing time stamp 1 by 20) because the switch was initially off and \emph{nothing else changed that} in the whole interval.
Another important feature that helps avoiding irrelevant information is that a negative literal \mbox{\tt not p} is never used in derivations, since it is understood as ``there is no cause for {\tt p}.''
If this was not done in this way, then the explanation for {\tt h(light,off,20)} in an answer set where no real action occurred would include the negation of {\em all combinations of actions} that could have changed the light value along the way from 1 to 20 (and there are too many, even in this simple example).
This information may be relevant for answering why was not the light turned on\footnote{Including ``why not'' queries is an interesting topic for future study.}, but is irrelevant for explaining why it has simply remained off, which is the purpose of \xclingo{}.

For adding custom trace labels to a program, the programmer has to make use of the \xclingo{}'s markup language. It works by adding annotations to the program that start with \texttt{\%!}, so they are just treated as comments by a plain ASP solver like \texttt{clingo}. There exist two different types of annotations for writing custom trace labels.
The first, \tracerule{}, allows the user to write a custom trace label and to associate it with a specific rule in the program. Listing~\ref{listing:f1} shows how we have modified lines 11--23 from Listing~\ref{listing:p1b} to associate some custom trace labels with those rules.

\begin{figure}[htbp]
\begin{lstlisting}[caption=Adding trace labels to specific rules with \tracerule{}., label={listing:f1}]
%%%%%% Indirect effects
%!trace_rule {"The relay is working at %",J}
  c(relay,on,J)   :- h(s1,closed,J), h(ab(relay),false,J), time(J).

%!trace_rule {"The relay is not working at %",J}
  c(relay,off,J)  :- h(s1,open,J), time(J).

%!trace_rule {"The relay is not working at %",J}
  c(relay,off,J)  :- h(ab(relay),true,J), time(J).

c(s2,closed,J)  :- h(relay,on,J), time(J).

%!trace_rule {"The light is on at %",J}
  c(light,on,J)    :- h(s2,closed,J), h(ab(bulb),false,J), time(J).

%!trace_rule {"The light is off at %",J}
  c(light,off,J)   :- h(s2,open,J), time(J).

%!trace_rule {"The light is off at %",J}
  c(light,off,J)   :- h(ab(bulb),true,J), time(J).

%%%%%% Malfunctioning
%!trace_rule {"The bulb has been damaged at %",I}
  c(ab(bulb),true,I) :- o(break,I), step(I).

%!trace_rule {"The relay has been damaged at %",I}
  c(ab(relay),true,I) :- o(surge,I), step(I).

%!trace_rule {"The bulb has been damaged at %",I}
  c(ab(bulb),true,I) :- o(surge,I), not h(b_prot,true,I-1), step(I).
\end{lstlisting}
\end{figure}
The \tracerule{} annotations are associated to the rule they precede. Inside the braces, the first argument is mandatory and it must be a string enclosed by quotes. The rest of the arguments are optional and must be variable names used either in the head or in the body of the rule. The \texttt{\%} placeholders are special characters that will be replaced by the values of the variables after solving, according to the order that variables are listed after the first argument.

Trying to write trace labels only using \tracerule{} annotations soon makes the code larger, redundant and harder to maintain. For those cases, \trace{} annotations are more suitable instead: they allow a permanent association of a label to an atom, regardless of which rules have triggered it.
Thus, their information is less specific but allows other general interesting features.
For instance, \trace{} annotations can be stored separately from the base code, multiple versions of the same trace labels could be written depending on the context: different languages, different users or different detail level. 
Lines 1--8 in Listing~\ref{listing:f2} show the \trace{} annotations added for obtaining the output from Fig.~\ref{fig:p1expl}.
For the part between braces, the syntax works the same as in \tracerule{} annotations, but instead of being followed by a rule, they are followed by a \textit{conditional atom} defining the set of atoms affected by the trace label. 
Finally, the \showtrace{} annotations (Lines 10--11 in Listing~\ref{listing:f2}) work in a similar way to the {\tt \#show} directives in {\tt clingo}, choosing which atoms are displayed in each answer set, but in this case, asking for their explanation.
Again, \showtrace{} annotations allow conditional atoms, as happened with \trace{}.

\begin{figure}[htbp]
\begin{lstlisting}[caption={Tracing atoms through \trace{} annotations for Program $P_1$.}, label={listing:f2}]
%!trace {"Hypothesis: there has been a power surge at %",J} o(surge,J).
%!trace {"Hypothesis: something has broken the bulb at %",J} o(break,J).
%!trace {"The agent has closed switch s1 at %",J} o(close(s1),J).

%!trace {"The % was initially damaged",C} h(ab(C),true,0).
%!trace {"Initially, the % was not damaged",C} h(ab(C),false,0).

%!trace {"% was initially %",F,V} h(F,V,0) : not abfluent(F).

%!show_trace h(light,V,1).
%!show_trace h(relay,V,1).
\end{lstlisting}
\end{figure}



\section{Implementation}
\label{sec:translation}

%
The tool \xclingo{} performs two main tasks: (1) a \emph{translation} of the annotated program $P$ into a logic program $P'$; and (2) a \emph{construction} of derivation trees by decoding the answer sets of $P'$.
Program $P'$ built in the translation phase is equivalent to the non-annotated version of $P$ but includes auxiliary predicates and {\tt clingo} theory atoms to keep track of the rules that have been fired.
The translation is further divided into two steps.
In a first step, \tracerule{} and \trace{} annotations become {\tt clingo} theory atoms without much transformation.
These atoms are accepted by the grounder and can be handled after solving through the {\tt clingo} Python API.
The \showtrace{} annotations are just transformed into traditional rules for an auxiliary head predicate \texttt{show\_all\_p} for each predicate {\tt p} to be shown.

Listing~\ref{listing:f4-phase1} shows the result of this first step when applied to some of the annotations presented before.

\begin{figure}[htbp]
\begin{lstlisting}[caption={Transforming annotations into theory atoms and auxiliary predicates.}, label={listing:f4-phase1}]
%% Translation of lines 2,3 in Listing 3
c(relay,on,J) :- h(s1,closed,J), h(ab(relay),false,J), time(J), 
                 &trace{"The relay is working at %",J}.

%% Translation of line 1 in Listing 4
&trace_all{o(surge,J),"Hypothesis: there has been a power surge at %",J : } :- o(surge,J).

%% Translation of line 10 in Listing 4
show_all_h(light,V,J):-h(light,V,J).
\end{lstlisting}
\end{figure}

In a second step, each predicate of the original program $P$ is prefixed with {\tt holds\_} and it is assigned a numeric identifier {\tt N}.
After that, each rule of the form {\tt H :- B, \&trace\{label\}} is split into the following rules:
\begin{eqnarray*}
\mathtt{fired\_N(X}_1,\dots,\mathtt{X}_n\mathtt{)} & \text{\tt :-} & \mathtt{B}\\
\mathtt{H} & \text{\tt :-} & \mathtt{fired\_N(X}_1,\dots,\mathtt{X}_n\mathtt{)}\\
\mathtt{\&trace\{N,H,label\}} & \text{\tt :-} & \mathtt{fired\_N(X}_1,\dots,\mathtt{X}_n\mathtt{)}
\end{eqnarray*}
where the {\tt X}$_1$, \dots, {\tt X}$_n$ also include the free variables in the body {\tt B}.
If the original rule body does not contain any trace label, then the last rule is not generated.
As an example, suppose that rule in lines 2--3 from Listing~\ref{listing:f4-phase1} is assigned identifier 33.
Then, its translation is shown in Listing~\ref{listing:fired}.
\begin{figure}[htbp]
\begin{lstlisting}[caption=Translation of lines 2--3 from Listing~\ref{listing:f4-phase1}., label={listing:fired}]
fired_33(relay,on,J) :-
     holds_h(s1,closed,J),holds_h(ab(relay),false,J),holds_time(J).
holds_c(Aux0,Aux1,Aux2) :- 
     fired_33(Aux0,Aux1,Aux2).
&trace{33,c(relay,on,J),"The relay is working at %",J } :-
     fired_33(relay,on,J).
\end{lstlisting}
\end{figure}

During this translation process, some additional rule information is stored: (1) the original head of the rule; (2) the original body of the rule; and (3) a list with all the variable names used in the \texttt{fired\_} rule.
This information is used to reconstruct the derivation proof of the atoms after solving.


Once the translated program $P'$ is generated, \xclingo{} makes a call to {\tt clingo}'s {\tt solve} function to retrieve its answer sets and, for each one, proceeds to construct the explanations from the information retrieved in the answer set. 
To do so, the first step consists in collecting all the theory atoms \texttt{\&trace} and \texttt{\&trace\_all} and replacing the \texttt{\%} placeholders in strings by their actual values. 
Once processed, the trace labels are stored into a dictionary, where they can be retrieved either by their associated \texttt{fired\_id} or by their atom.
Then, \xclingo{} identifies which rules have been fired for each model by finding the \texttt{fired\_} atoms in it. For each fired rule, we save the different sets of values the rule was fired with in a dictionary indexed by \texttt{fired\_id}. 
With this information, together with the information about the original rules that was stored during translation (the original head, the original body, and all the variable names) we build the derivation proofs and print the explanations.
The construction of the derivation proof is made with a ``\textit{causes table}'' 
%
that has a row per each rule and includes, the trace labels and the bodies that fired those labels.
Once the \texttt{causes table} is obtained, the next step is filtering those atoms affected by \showtrace{} annotations. All the atoms in the model that start with the \texttt{show\_all\_} prefix are retrieved and stored in a list after removing the prefix. If that list is empty, then all the atoms in the model are explained.
Finally, the explanation of each atom in that list has to be built and printed. Listing~\ref{listing:build-explanations-pseudocode} shows the pseudocode for the recursive function that builds the explanation graph for a given atom and a given causes table.
According to the tree structure, each explanation is a python dictionary where keys are trace labels and values are in another dictionary (a subtree).
Since an atom can have multiple explanations, the function returns a list of dictionaries.
For a leaf of the tree graph (a fact), the result is a list with an empty dictionary (Line 12).
\begin{figure}[htbp]
\begin{lstlisting}[basicstyle=\small, language=Python, caption=Pseudocode for the  \texttt{build\_explanations} function., label={listing:build-explanations-pseudocode}]
def build_explanations(atom, causes_table, stack):
    for row in causes_table.find_by_fired_head(atom):
        if not_empty(row['f_body']):
            entry_expls = []  # Explanations of the current row.
            for atom a in row['f_body']:
                if atom a not in stack:
                    stack.push(a)
                    atom_expls = build_explanations(a, causes, stack)
                    stack.pop(a)
                    entry_expls = _combine(entry_expls, a_expls)
        else:
            entry_expls = [{}]

        if not_empty(row['traces']):
            explanations = []
            for t in row['traces']:
                for e in entry_expls:
                    explanations.append({t: e})
        else:  # skip nodes without trace labels
            explanations = entry_expls

    return explanations
\end{lstlisting}
\end{figure}
While exploring the explanations of an atom $A$, we can find that one atom $a$ in its fired body has multiple explanations. In that case, $A$ has a different explanation for each $a$ explanation. Function \texttt{\_combine} in line 10, performs this combination.
Lines 14 to 21 manage the trace labels, which are the root of the tree graphs. The atom has one explanation for each different trace label it is associated with. If an atom has no trace labels, nothing is added as root of the explanations (lines 20--21). As a consequence, one level is skipped in that subtree.

Given that atoms may have multiple alternative explanations, the size of the set of explanations, when expressed as a set of trees, can grow exponentially in the worst case.
To see why, just consider the program in Listing~\ref{listing:chain} for some constant value $n$.
Here, the explanation for atom {\tt p($n$)} is a chain formed by $n$ labels {\tt "a($n$)"}, \dots, {\tt "a(1)"}.
If we add a second trace for {\tt p(X)} using  {\tt "b(\%)"}, then each atom {\tt p(X)} has now two alternative labels {\tt a(X)} and {\tt b(X)}, and so, the number of alternative derivations for {\tt p($n$)} becomes $2^n$.
\begin{figure}[htbp]
\begin{lstlisting}[caption=A sequential chain for predicate {\tt p}.,label=listing:chain]   
p(1).
p(X+1) :- p(X), X<=n.
%!trace {"a(%)",X} p(X).
\end{lstlisting}
\end{figure}

\begin{figure}[]
\centering
\begin{minipage}{.475\textwidth}
  \includegraphics[width=1\linewidth]{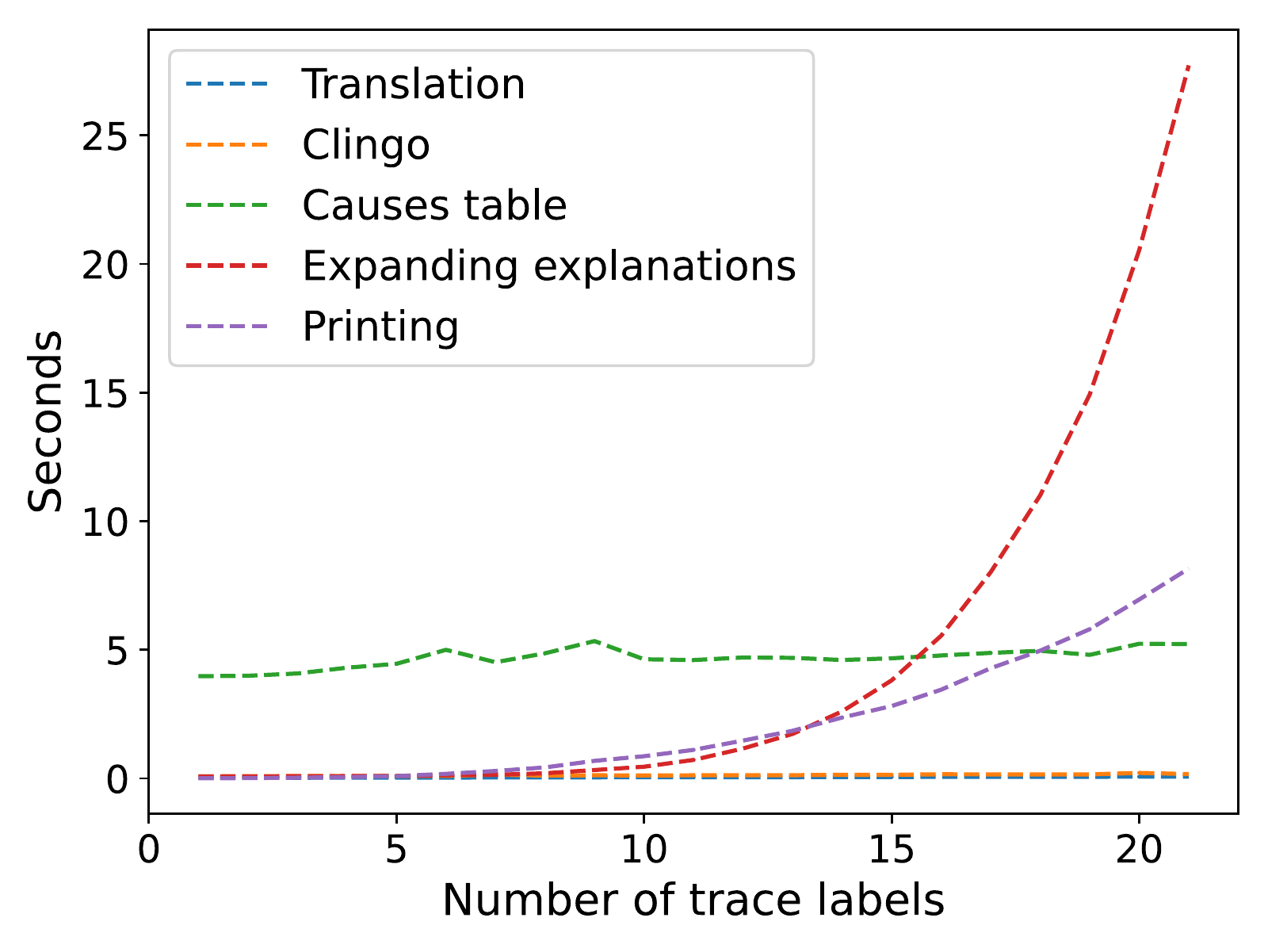}
  \caption{Execution time vs labels per atom.}
  \label{fig:times-traces}
\end{minipage}%
\hfill
\begin{minipage}{.475\textwidth}
  \includegraphics[width=1\linewidth]{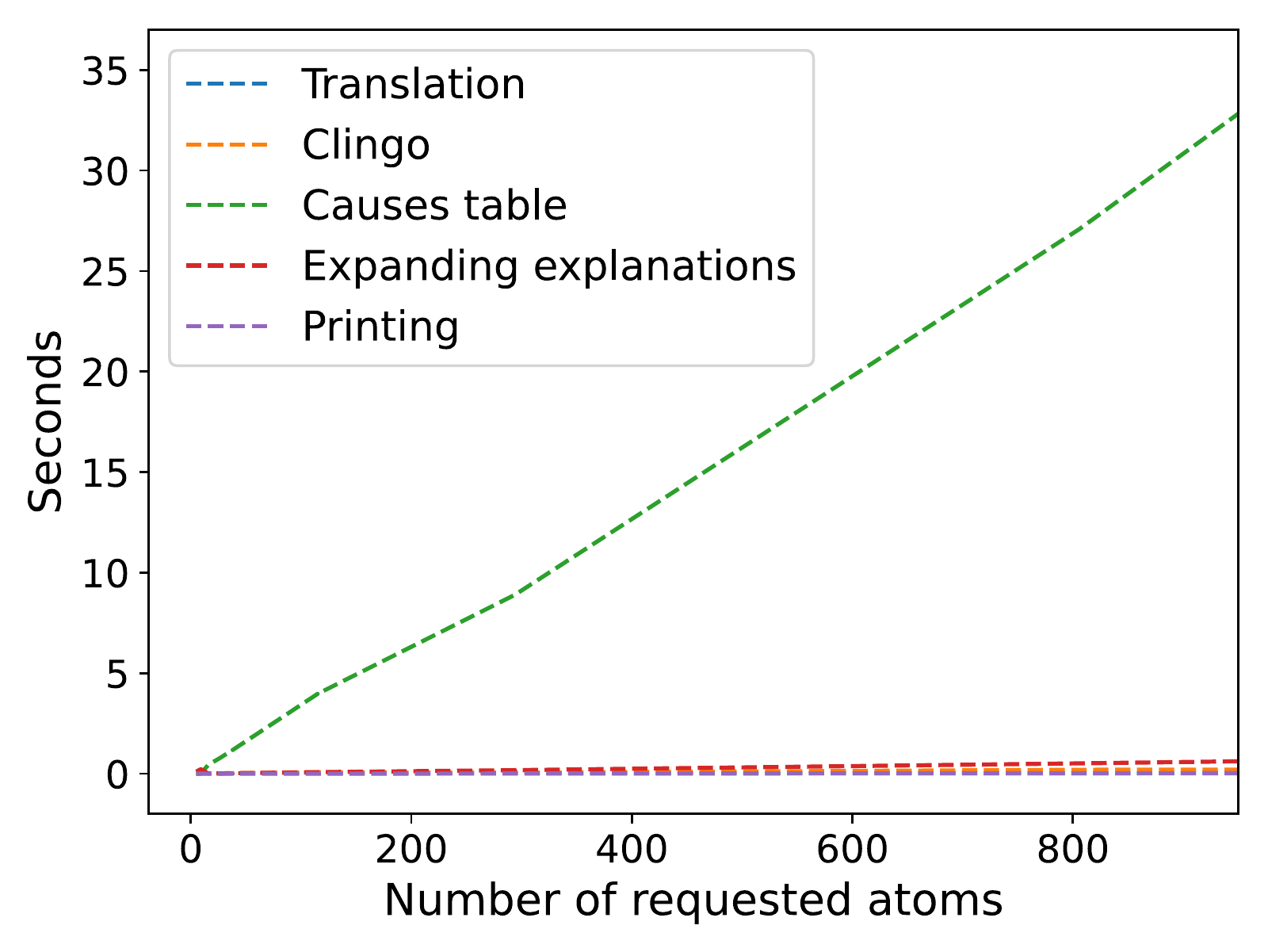}
  \caption{Execution time vs num. of atoms.}
  \label{fig:times-atoms}
\end{minipage}
\end{figure}

We have tested the performance of \xclingo{} on a simple encoding of the well-known \textit{blocks world} domain varying the size of the scenario in terms of the number of blocks, but also the size of the explanations, by adding more trace labels per atom.
We asked \xclingo{} to explain the actions performed, the final location of each block and the final value of predicate {\tt unclear(B)} that states whether a block {\tt B} has anything on top (see example in Fig.~\ref{fig:block-explanation}).
\begin{figure}[]
\begin{lstlisting}
>> unclear(9,14)        [1]
  *
  |__"Block 9 is finally unclear"
  |  |__"Block 1 is finally on 9"
  |  |  |__"Block 1 was moved on top of block 9 at t=11"
\end{lstlisting}
\caption{Explanation answering why block 9 is unclear at final step 14.}
\label{fig:block-explanation}
\end{figure}
We measured the time spent in five different steps of the process: (1) the translation; (2) the execution of {\tt clingo} for solving the planning problem; (3) the construction of the causes table and the dependencies among labels; (4) the expansion of all derivation trees; and finally, (5) their printing.
Fig.~\ref{fig:times-traces} shows how the different times evolve when we fix the number of atoms requested, but we progressively add new trace labels per each atom.
As we explained before, increasing the number of labels per atom causes an exponential grow in the number of alternative explanations that is well reflected in the graphic.
As future work, we plan to include an execution mode for generating a maximum number of explanations per atom instead of expanding all of them.
Fig.~\ref{fig:times-atoms} shows how, when we fix the number of trace labels per atom, the time spent in the construction of the causes table and its dependencies has a linear increase as we add more atoms.
We also plan to explore alternatives to the construction of the explanations like performing a tabled evaluation or even using {\tt clingo} itself to solve this part of the problem.

\section{Related Work}
\label{sec:related-work}

The explanations of \xclingo{} correspond to \textit{causal justifications} from \cite{CabalarFF14cg}
except that \xclingo{} does not guarantee that the derivation trees are minimal, although it performs some minor simplifications.
%
In the survey \cite{FandinnoS19}, different approaches to the problem of answering the ``why" in ASP are reviewed, including the two related ones called \textit{off-line justifications} \cite{pontelli2009justifications} and \textit{argumentative explanations} \cite{schulzT2016justifying}. 
An important difference with respect to these approaches is that, as we explained, \xclingo{} does not derive information from negative literals. This is because default negation is understood, under \xclingo{} semantics, as the absence of cause.
As a result, explanations only provide information when defaults are broken (like inertia), rather than including all possible events that could have changed the outcome, although they did not happen.
The latter can be interesting for answering ``why not'' queries, but the size of explanations may simply become unmanageable.
Another difference is that, thanks to the labelling system inherited from causal justifications, \xclingo{} allows selecting in a fine grained way which information can be used in the justification trees.

Other systems that provide explanations for logic programs are  \texttt{ErgoAI}\footnote{\url{http://coherentknowledge.com/product-overview-ergoai-platform/}}, for Datalog programs, and {\tt s(ASP)}~\cite{AriasCSM18} for non-ground ASP programs.
An advantage of these two systems is that their explanations can be generated without resorting to a full grounding of the program.
Besides, \texttt{ErgoAI} shows some similarities to \xclingo{} like selecting the information in the explanations (although, in this case, we must explicitly define which information must be omitted), or allowing text labels and variables in the justifications.
However, both \texttt{ErgoAI} and {\tt s(ASP)} include information about negated literals, unlike \xclingo{}.

\section{Conclusions and Future Work}
\label{sec:conc}


We presented \xclingo{}, an ASP extension interpreter that allows obtaining justifications of the literals in the answer sets of logic programs. At the moment, the tool is a partial implementation of the Logic Programming extension described by \textit{causal justifications}~\cite{CabalarFF14cg}. 
The python source code of \xclingo{} and instructions for its usage are available at github\footnote{\url{https://github.com/bramucas/xclingo}}. It requires python 3.* and the following python modules to be installed: clingo\footnote{\url{https://potassco.org/clingo/python-api/5.4/}}, pandas\footnote{\url{https://pandas.pydata.org/}} and more\_itertools\footnote{\url{https://github.com/more-itertools/more-itertools}}. 
We illustrated, on a diagnostic reasoning example, how \xclingo{}'s features can be used for obtaining readable explanations, that can be even expressed in natural language. Even if explanations are not needed, \xclingo{} can also be used as a complement to \texttt{clingo} for debugging purposes, since the annotations are included as comments and do not make the code incompatible.


The current version of \xclingo{} is still in a preliminary stage and will be augmented with new features. Immediate future work includes the treatment of minimization directives (something essential, for instance, to obtain minimal solutions for diagnosis problems), and other usual {\tt clingo} constructs like \textit{choice rules} or \textit{pooling}, which are not accepted yet.
Another extension we plan to include in the future is the addition of trace labels for constraints, so that those that are labelled become weak constraints. Other extensions for the long term include the use of causal literals as in~\cite{Fan16} or the use of verification annotations to be combined with a similar theorem proving technique as done in~\cite{lifschitz2019anthem}.


\nocite{*}
\bibliographystyle{eptcs}
\bibliography{bibliography}
\end{document}